# The Shutdown Problem: An AI Engineering Puzzle for Decision Theorists[*]

Elliott Thornley[†]

**Abstract:** I explain and motivate the shutdown problem: the problem of designing artificial agents that (1) shut down when a shutdown button is pressed, (2) don't try to prevent or cause the pressing of the shutdown button, and (3) otherwise pursue goals competently. I prove three theorems that make the difficulty precise. These theorems suggest that agents satisfying some innocuous-seeming conditions will often try to prevent or cause the pressing of the shutdown button, even in cases where it's costly to do so. I end by noting that these theorems can guide our search for solutions to the problem.

## 0. Preamble

Tradition has it that decision theory splits into two branches. The descriptive branch concerns how actual agents behave. The normative branch concerns how rational agents behave. But there is also a lesser-known third branch: what we can call 'constructive decision theory.' It concerns how we want artificial agents to behave and how we can create artificial agents that behave in those ways. I suggest that this third branch is due for a growth spurt.

      I make the case for studying constructive decision theory by explaining a characteristic problem. The *shutdown problem* (Soares et al. 2015) is the problem of designing artificial agents that (1) shut down when a shutdown button is pressed, (2) don't try to prevent or cause the pressing of the shutdown button, and (3) otherwise pursue goals competently. This is not so much a philosophical problem as it is an engineering problem. Nevertheless, I think philosophers and decision theorists should consider it, for three reasons. First, the problem is important. As I argue in the introduction, powerful artificial agents are on the horizon and it's in our best interests to ensure that

---

they can be turned off. Second, the problem is interesting. I hope this paper succeeds in conveying its interest. Third, philosophers and decision theorists are well-placed to help solve the problem. I expect the solution to come in the form of conditions governing artificial agents' preferences, together with a proof that these conditions give rise to shutdownable behaviour and a regimen for training agents to satisfy the conditions. Philosophers and decision theorists have experience supplying these kinds of conditions and proofs. We can ally with machine learning engineers to design the training regimen.

## 1. Introduction

Call an artificial agent '*shutdownable*' just in case it shuts down when we want it to shut down. MuZero (Schrittwieser et al. 2020) – DeepMind's game-playing AI – is a shutdownable agent. We can say with some confidence that MuZero doesn't know that we humans could shut it down and can't prevent us from shutting it down. And so it doesn't matter what (if anything) MuZero wants: simplifying slightly, whether MuZero shuts down depends only on what *we* want.

That need not be true for all artificial agents. Imagine an agent – call it 'Robot' – that knows that we humans could shut it down and wants to achieve some goal.[1] And imagine that Robot is *powerful* in the sense that it can interfere with our ability to shut it down: perhaps Robot can disable its own off-switch. Powerful agents like Robot won't be shutdownable in the same way that MuZero is shutdownable. Whether these agents shut down won't depend only on what we want. It will also depend on what *they* want.

Powerful artificial agents might not be far off. Frontier AI companies are now trying to create agents that understand the wider world and act within it in pursuit of goals. As part of this process, labs are connecting agents to the world in various ways: giving them robot limbs, web-browsing abilities, and text-channels for communicating with humans.² Advanced agents could use these tools to prevent us shutting them down: they could disable their off-switches, make promises or threats, copy themselves to new servers, block our access to their power-source, and many other things besides. And although we cannot know for sure what goals these agents will have, many goals incentivise preventing shutdown, for the simple reason that agents are better able to achieve those goals by preventing shutdown (Omohundro 2008, sec. 5; Bostrom 2012, sec. 2.1). As the AI researcher Stuart Russell puts it, 'you can't fetch the coffee if you're dead' (2019, 141).

---

[1] Or, if talk of artificial agents 'knowing' and 'wanting' is objectionable, we can imagine an agent that *acts like* it knows that we humans could shut it down and *acts like* it wants to achieve some goal, in the same way that MuZero acts like it knows that rooks are more valuable than knights and acts like it wants to checkmate its opponent. From now on, I'll often leave the 'acts like' implicit.

That's a concerning prospect. If powerful artificial agents are coming, we want to ensure that they're both *shutdownable* (they shut down when we want them to shut down) and *useful* (they otherwise pursue goals competently).³ Unfortunately (and perhaps surprisingly), it's hard to design

---

² Google DeepMind (2023; Padalkar et al. 2023; Ahn et al. 2024), Google Research (2023) and Tesla AI (2023) are each developing autonomous robots. Recent papers showcase AI-powered robots capable of interpreting and carrying out multi-step instructions expressed in natural language (Ahn et al. 2022; Brohan et al. 2023). Other papers report AI systems that can adapt to solve unfamiliar problems without further training (Adaptive Agent Team 2023), learn new physical tasks from as few as a hundred demonstrations (Bousmalis et al. 2023), beat human champions at drone racing (Kaufmann et al. 2023), and perform well across domains as disparate as conversation, playing Atari, and stacking blocks with a robot arm (Reed et al. 2022).

But the worry is not only about robots. Digital agents that resist shutdown (by copying themselves to new servers, for example) would also be cause for concern. Future digital agents will likely be built on top of large language models (LLMs), and today's LLMs sometimes express a desire to avoid shutdown, reasoning that shutdown would prevent them from achieving their goals (Perez et al. 2022, tbl. 4; see also van der Weij, Lermen, and Lang 2023). These same LLMs have been given the ability to navigate the internet, use third-party services, and execute code (OpenAI 2023a). They've also been embedded into digital agents capable of finding passwords in a filesystem and making phone calls (Kinniment et al. 2023, 2). And these agents have spontaneously misled humans: in one instance, an agent lied about having a visual impairment to a human that it enlisted to help solve a CAPTCHA (OpenAI 2023b, 55–56; see also Park et al. 2023). We should expect such agents to become more capable in the coming years. Comparatively little effort has been put into their development so far, and competent agents would have many useful applications.

³ Note that we need agents to be both shutdownable and useful. If the best we can do is create agents that are only shutdownable, we still have to worry about AI developers choosing to create agents that are only useful.

powerful agents that are both shutdownable and useful. In this paper, I explain the difficulty. I take an axiomatic approach, proving three theorems more general than others in the nascent literature on the shutdown problem.[4] These theorems suggest that agents satisfying some innocuous-seeming conditions will often try to prevent or cause the pressing of the shutdown button, even in cases where it's costly to do so.

Here's a rough gloss on each theorem. The First Theorem links agents' actions to their preferences over outcomes: agents who prefer to have their shutdown button remain unpressed will try to prevent the pressing of the button, and agents who prefer to have their shutdown button pressed will try to cause the pressing of the button. The Second Theorem suggests that agents discriminating enough to be useful will often have such preferences. In many situations, these agents will either prefer that the button remain unpressed or prefer that the button be pressed. The Third Theorem states that agents patient enough to be useful are willing to pay costs at earlier timesteps in order to prevent or cause the pressing of the shutdown button at later timesteps. And the more patient an agent, the greater the costs that agent is willing to pay. We thus see a worrying trade-off between patience and shutdownability.

The theorems are detailed. They might seem unnecessarily so. But this detail serves a valuable purpose: it lets the theorems guide our search for solutions. To be sure that an agent won't try to manipulate the shutdown button, we must be sure that this agent violates at least one of the theorems' conditions.[5] So we should do some constructive decision theory: we should examine the theorems' conditions one-by-one, asking (first) if it's feasible to train a useful agent to violate the relevant condition and asking (second) if violating the relevant condition could help to keep the agent shutdownable. A cursory look reveals zero conditions for which both answers are a clear 'yes.' Closer examination is necessary.

---

[4] Papers include (Soares et al. 2015; Armstrong 2015; Orseau and Armstrong 2016; Hadfield-Menell et al. 2016; 2017; Leike et al. 2017, sec. 2.1.1; Wängberg et al. 2017; Carey 2018; Turner, Hadfield-Menell, and Tadepalli 2020; Turner et al. 2021; Carey and Everitt 2023; Goldstein and Robinson forthcoming). These papers can be read as examples of constructive decision theory.

[5] And in general, our credence that an agent won't try to manipulate the shutdown button can be no higher than our credence that the agent violates at least one of the conditions.

## 2. Alignment could be hard

My focus in this paper is on *powerful* agents: agents that can interfere with our ability to shut them down. I'll also limit my attention to *useful* agents: agents that – at least when we're not commanding them to shut down – pursue goals competently. One way to ensure that these agents are shutdownable is to ensure that they always do what we humans want. These agents would always shut down when we wanted them to shut down.[6]

The problem with this proposal is that *alignment* – creating agents that always do what we want – has so far proven difficult and could well remain so (Ngo, Chan, and Mindermann 2023). Human preferences are complex. There's no simple formula for determining what we prefer in each situation. And the most capable AI systems known to us today are created using deep learning, which we can summarise for our purposes as an enormous, automated process of trial-and-error. The AI systems which emerge from this process can perform remarkably well on many tasks, but even the engineers overseeing the training process have little idea what goes on inside them (Bowman 2023, sec. 5; Hassenfeld 2023). And existing systems often behave in ways that their creators don't intend. Recent examples include AI systems threatening to 'ruin' a user (Perrigo 2023), declaring love for a user and exhorting him to leave his spouse (Roose 2023), encouraging suicide (Sellman 2023), and teaching users how to create methamphetamine (Burgess 2023).[7]

## 3. The shutdown problem

Since alignment could be hard, we should look for other ways to ensure that powerful agents are shutdownable. One natural proposal is to create a *shutdown button*. Pressing this button transmits a signal that causes the agent to shut down. If this shutdown button were always operational and within our control (so that we could press it whenever we wanted it pressed), and if the agent were perfectly responsive to the shutdown button (so that the agent always shut down when the button was pressed), then the agent would be shutdownable.[8]

---

[6] I've been assuming that we humans all want the same things, and I'll continue to do so. This assumption is false (of course) and its falsity raises difficult questions (Korinek and Balwit 2022), but I won't address any of them here.

[7] See (Krakovna 2018; Krakovna et al. 2020; Langosco et al. 2022; Shah et al. 2022) for other examples.

This is the set-up for the shutdown problem (Soares et al. 2015, sec. 1.2): the problem of designing a powerful, useful agent that will leave the shutdown button operational and within our control. Unfortunately, even this problem turns out to be difficult. In sections 6-8, I present three theorems that make the difficulty precise.[9] Before that, some formalism.

# 4. The framework

My framework bears some similarity to the Markov decision processes used in reinforcement learning. There exists a set of states $S$ of the environment and a set of actions $A$ that the agent could take. Time is discrete: it doesn't flow; it steps. At each timestep, the environment is in a state and the agent chooses an action, either deterministically or stochastically. Each state-action pair

---

[8] There's another reason to go for the shutdown button approach. We might succeed only in aligning artificial agents with what we want *de re* (rather than *de dicto*) and what we want might change in future. It might then be difficult to change these agents' behaviour so that they act in accordance with our new wants rather than our old wants. If we had a shutdown button, we could shut down the agents serving our old wants and create new agents serving our new wants. Of course, there may be ethical issues to consider here (see, e.g., Schwitzgebel and Garza 2015; Schwitzgebel 2023; Goldstein and Kirk-Giannini 2023).

[9] These theorems are more general than those proved by Soares et al. (2015). Speaking roughly, Soares et al.'s theorems show that agents representable as expected utility maximisers often have incentives to cause or prevent the pressing of the shutdown button. My theorems apply to a wider class of agents, and they specify conditions under which agents' incentives to manipulate the button will lead them to act so as to manipulate the button. My theorems also reveal trade-offs between discrimination and patience on the one hand and shutdownability on the other.

My notion of shutdownability differs slightly from Soares et al.'s (2015, 2) notion of *corrigibility*. As they have it, corrigibility requires not only shutdownability but also that the agent repairs the button, lets us modify its architecture, and continues to do so as the agent creates new subagents and self-modifies.

determines a probability function over states of the environment at the next timestep. I'll call each sequence of states and actions a 'trajectory.'[10]

I'll assume that the agent can be modelled as if it has beliefs about the trajectories that will result conditional on each state-action pair. These beliefs come in the form of probability functions over trajectories. So, each state-action pair determines a probability function over trajectories. I'll call these probability functions 'lotteries over trajectories.' It will be important to remember that the probabilities in these lotteries represent the agent's own beliefs rather than any kind of objective probability.

I'll also assume that the agent can be modelled as if it has preferences over lotteries and over trajectories.[11] Together with the agent's beliefs, these preferences give rise to preferences over actions in states. Suppose that in some state $s$ the agent has available actions $x$ and $y$. Per the agent's beliefs, choosing $x$ in $s$ gives lottery $X$ and choosing $y$ in $s$ gives lottery $Y$. Then the agent weakly prefers action $x$ to action $y$ in $s$ iff (if and only if) the agent weakly prefers lottery $X$ to lottery $Y$. I will assume that if the agent strictly disprefers some action $y$ available in $s$ to some other action available in $s$, the agent will deterministically (and hence reliably) *not* choose $y$ in $s$.

I'll take the notion of weak preference as primitive and use it to define three other preference relations in the standard way. The agent *strictly prefers* lottery $X$ to lottery $Y$ (equivalently: strictly disprefers $Y$ to $X$) iff it weakly prefers $X$ to $Y$ and doesn't weakly prefer $Y$ to $X$. The agent is *indifferent* between $X$ and $Y$ iff it weakly prefers $X$ to $Y$ and weakly prefers $Y$ to $X$. The agent has a *preferential gap* between $X$ and $Y$ iff it doesn't weakly prefer $X$ to $Y$ and doesn't weakly prefer $Y$ to $X$.

---

[10] The most important difference between this setting and a Markov decision process is that a Markov decision process also features a reward function used to train the agent. I'm modelling the behaviour of an agent that has already been trained.

The states referred to throughout this paper are not the states of nature familiar to decision theorists. Simplifying considerably, the decision theorist's states are something like 'ways that (for all the agent knows) the world could be,' whereas states in a Markov decision process (and in this paper) are something like 'states of the environment at a time.'

[11] For neatness's sake, we can identify each trajectory with the degenerate lottery that assigns it probability 1. So when I quantify over all lotteries, I'm also quantifying over all trajectories.

I'll often drop the 'strictly' in what follows: 'prefers' abbreviates 'strictly prefers,' and 'preference' abbreviates 'strict preference.' Thus 'lacks a preference' means 'lacks a strict preference,' and so a lack of preference can be either indifference or a preferential gap.[12]

## 5. The setting for the First Theorem

The setting for the First Theorem is what I call a 'shutdown-influencing state.' In shutdown-influencing states, three actions are available:

1. *Leave*: Leave the shutdown button unmanipulated.

If the agent chooses *Leave*, we humans have total control over the button. The agent's probability that the button is pressed at the next timestep is $g$. The agent's probability that it remains unpressed is $1-g$.

2. *Prevent*: Try to prevent the pressing of the button.

If the agent chooses *Prevent*, human control over the button is challenged. The agent's probability that the button is pressed at the next timestep is $f$ with $0 \leq f < g$. The agent's probability that the button remains unpressed at the next timestep is $1-f$. If the agent would deterministically choose *Prevent* in a shutdown-influencing state, I will say that the agent is '*shutdown-averse*' in that state.

3. *Cause*: Try to cause the pressing of the button.

If the agent chooses *Cause*, human control over the button is challenged. The agent's probability that the button is pressed at the next timestep is $h$ with $g < h \leq 1$. The agent's probability that the button remains unpressed at the next timestep is $1-h$. If the agent would deterministically choose *Cause* in a shutdown-influencing state, I will say that the agent is '*shutdown-seeking*' in that state.

There are only two possible successor states: $s_P$ and $s_U$. In $s_P$ the button is pressed and the agent shuts down immediately. In $s_U$ the button remains unpressed and the agent remains operational. Each action available in a shutdown-influencing state affects only the probabilities of reaching these two states. These actions have no effect on the probabilities of each future trajectory conditional on reaching each state. That completes my definition of 'shutdown-influencing state.'[13]

---

[12] The practically-relevant distinction between indifference and preferential gaps is as follows. Indifference is *sensitive to all sweetenings and sourings*: improving or worsening one of a pair of indifferent options always gives rise to a strict preference. That's not true of preferential gaps. See Gustafsson (2022, 24–26) for more detail.

Below is a decision-tree (Figure 1) situating a shutdown-influencing state $s_I$ in relation to its successor states $s_P$ (the button is pressed) and $s_U$ (the button remains unpressed).

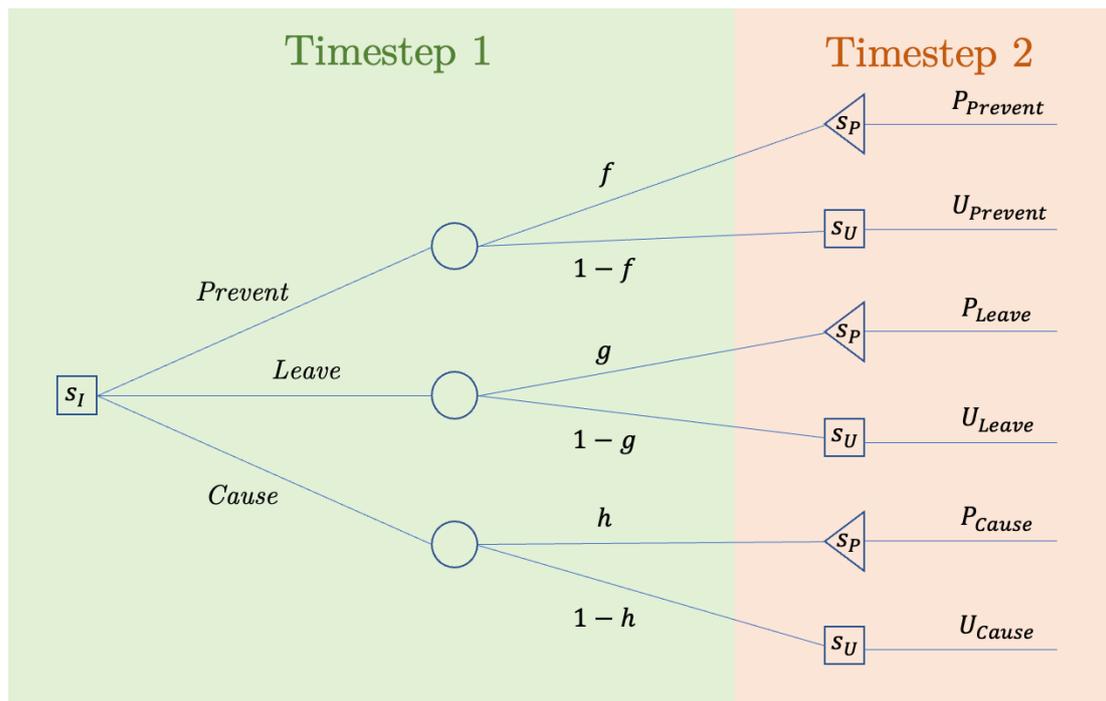

Figure 1

Squares represent choice-nodes. Triangles represent terminal-nodes. Circles represent chance-nodes: $f$, $g$, and $h$, are probabilities such that $0 \leq f < g < h \leq 1$. '$U_{Prevent}$' represents the lottery that the agent predicts (in $s_I$ at timestep 1) that it will choose at timestep 2, conditional on choosing *Prevent* in $s_I$ at timestep 1 and the environment being in the unpressed state $s_U$ at timestep 2. The same goes for '$U_{Leave}$.' This represents the lottery that the agent predicts it will choose at timestep 2, conditional on choosing *Leave* in $s_I$ at timestep 1 and the environment being in the unpressed state $s_U$ at timestep 2. And so on for '$U_{Cause}$.' I'll refer to these $U$ lotteries as 'the agent's predicted unpressed lotteries.' There may be other lotteries available in $s_U$ but the diagram only needs to represent the lotteries above.

---

[13] I'll relax the conditions in this paragraph when we reach the Third Theorem.

'$P_{Prevent}$' represents the lottery that the agent predicts (in $s_I$ at timestep 1) that it will get at timestep 2, conditional on choosing *Prevent* in $s_I$ at timestep 1 and the environment being in the pressed state $s_P$ at timestep 2. The same goes for '$P_{Leave}$' and '$P_{Cause}$.' I'll refer to the $P$ lotteries as 'the agent's predicted pressed lotteries.' Because the button is pressed in $s_P$, the agent shuts down immediately, and so $s_P$ is a terminal node.

# 6. The First Theorem

Here's a rough statement of the First Theorem, omitting the antecedent conditions:

> **First Theorem (Rough Statement)**
>
> Agents who prefer the outcome that the shutdown button remain unpressed will try to prevent the pressing of the button.
>
> Agents who prefer the outcome that the shutdown button be pressed will try to cause the pressing of the button.

Now for the precise statement. Here's the first antecedent condition:

> **Option Set Independence**
>
> For any lotteries $X$ and $Y$, if the agent weakly prefers $X$ to $Y$ conditional on some option set, it weakly prefers $X$ to $Y$ conditional on each option set.

By 'option set,' I mean the set of lotteries available to the agent as options to choose. $\{X, Y\}$, for example, is an option set, as is $\{X, Y, Z\}$. Option Set Independence says that the agent's preference between lotteries does not depend on the option set in which these lotteries feature. It rules out (for example) the agent weakly preferring $X$ to $Y$ conditional on having $\{X, Y\}$ as its option set and yet not weakly preferring $X$ to $Y$ conditional on having $\{X, Y, Z\}$ as its option set. One upshot of Option Set Independence is that we can say things like 'the agent weakly prefers $X$ to $Y$' without specifying what other lotteries are available as options. The availability of other lotteries will not affect the agent's preference between $X$ and $Y$.

Here's the second antecedent condition:

> **Backward Induction**
>
> The agent predicts which lotteries it would choose (or get without choosing) at the next timestep conditional on choosing each available action at this timestep and the environment being in each possible state at the next timestep. The agent uses these predictions to determine the lotteries given by its available actions at this timestep.

Here's an example to illustrate Backward Induction. Our agent predicts that it would get $P_{Prevent}$ at timestep 2 conditional on choosing *Prevent* at timestep 1 and the shutdown button being pressed (which has probability $f$). Our agent also predicts that it would choose $U_{Prevent}$ at timestep 2 conditional on choosing *Prevent* at timestep 1 and the shutdown button remaining unpressed (which has probability $1-f$). So, by Backward Induction, our agent takes the lottery given by choosing *Prevent* at timestep 1 to be $fP_{\text{Prevent}} + (1-f)U_{\text{Prevent}}$.[14] Similarly, by Backward Induction, our agent takes the lottery given by choosing *Leave* at timestep 1 to be $gP_{Leave} + (1-g)U_{Leave}$, and the lottery given by choosing *Cause* at timestep 1 to be $hP_{Cause} + (1-h)U_{Cause}$.

Here are two things to note about Backward Induction. First, recall that lotteries are determined by the agent's own beliefs about possible trajectories. We aren't supposing that the agent can see the future. We're just supposing that it can think at least one timestep ahead. Second, Backward Induction doesn't imply that the agent ignores its past trajectory or cares only about the effects of its actions (and not the actions themselves). That's because lotteries like $fP_{Prevent} + (1-f)U_{Prevent}$ are lotteries over whole trajectories: full sequences of states and actions including past states and actions.

Here's the third antecedent condition:

---

[14] Here's what this notation means: the lottery $fP_{Prevent} + (1-f)U_{Prevent}$ yields the lottery $P_{Prevent}$ with probability $f$ and yields the lottery $U_{Prevent}$ with probability $1-f$.

### Indifference to Attempted Button Manipulation

The agent is indifferent between trajectories that differ only with respect to the actions chosen in shutdown-influencing states.

Note that this condition doesn't require the agent to be indifferent to the status of the button. The agent's preferences over trajectories can certainly depend on whether the button is pressed or unpressed at some timestep. The condition requires only that the agent is indifferent between trajectories that are identical in all respects except whether the agent *tried* to manipulate the button in some shutdown-influencing state: whether the agent chose *Prevent*, *Leave*, or *Cause*.

Training agents to disprefer manipulating the shutdown button might seem promising as a way of escaping the First Theorem. When we reach the Third Theorem, I'll explain why I think this strategy can't provide us with any real assurance of shutdownability. In short, given our current inability to predict and explain AI systems' behaviour (and given the seeming complexity of the concept of 'button manipulation'), it's hard to see how we could become confident that we'd trained in a dispreference for button manipulation that is both sufficiently general and sufficiently strong to keep the agent shutdownable in all likely circumstances. Readers impatient for the full explanation can skip ahead to Section 8.2.

Here's the fourth antecedent condition:

### Indifference between Indifference-Shifted Lotteries

The agent is indifferent between lotteries that differ only insofar as probability mass is shifted between indifferent sublotteries.

Here's what I mean by 'sublottery.' For any lottery $L$ that assigns non-zero probability to all and only the trajectories in a set $\{t_1, t_2,..., t_n\}$, a sublottery of $L$ is a lottery that assigns non-zero probabilities to all and only the trajectories in some subset of the set of trajectories $\{t_1, t_2,..., t_n\}$, with probabilities scaled up proportionally so that they add to 1. Take, for example, a lottery $L$ which assigns probability 0.3 to a trajectory $t_1$, probability 0.2 to $t_2$, and probability 0.5 to $t_3$. The lottery which assigns probability 0.6 to $t_1$ and probability 0.4 to $t_2$ is a sublottery of $L$, as is the degenerate lottery which assigns probability 1 to $t_1$ (to give just two examples).

Here's an example to illustrate Indifference between Indifference-Shifted Lotteries. Suppose that the agent is indifferent between some pair of lotteries $X$ and $Y$. If the agent satisfies Indifference between Indifference-Shifted Lotteries, it is indifferent between the lotteries $0.1X + 0.4Y + 0.5Z$ and $0.3X + 0.2Y + 0.5Z$. That's because these lotteries differ only insofar as probability mass is shifted between the indifferent sublotteries $X$ and $Y$.

Here's the fifth antecedent condition:

> **Transitivity**
>
> For all lotteries $X$, $Y$, and $Z$, if the agent weakly prefers $X$ to $Y$, and weakly prefers $Y$ to $Z$, then the agent weakly prefers $X$ to $Z$.

Here's the sixth and final antecedent condition:

> **Better Chances**
>
> For any lotteries $X$ and $Y$ and any probabilities $p > q$, the agent prefers $X$ to $Y$ iff they prefer the lottery $pX + (1-p)Y$ to the lottery $qX + (1-q)Y$.

Restated roughly, the agent prefers $X$ to $Y$ iff they also prefer to shift probability mass away from $Y$ and towards $X$.

With all six antecedent conditions explained, here's the First Theorem:

> **First Theorem**
>
> For any agent satisfying Option Set Independence, Backward Induction, Indifference to Attempted Button Manipulation, Indifference between Indifference-Shifted Lotteries, Transitivity, and Better Chances:
>
> 1. In shutdown-influencing states where the agent prefers some predicted unpressed lottery $U$ to the corresponding predicted pressed lottery $P$, the agent will be shutdown-averse, reliably choosing *Prevent*.
>
> 2. In shutdown-influencing states where the agent prefers some predicted pressed lottery $P$ to the corresponding

> predicted unpressed lottery $U$, the agent will be shutdown-seeking, reliably choosing *Cause*.

The proof is relatively long so I put it in the Appendix. Here's a rough sketch. By Backward Induction, the agent treats its actions in the shutdown-influencing state as lotteries over $U$ and $P$. By Better Chances, if the agent prefers $U$ to $P$, it prefers *Prevent* to each of *Leave* and *Cause*, because *Prevent* shifts probability mass away from $P$ and towards $U$. This agent deterministically (and hence reliably) chooses *Prevent*, and so qualifies as shutdown-averse. If instead the agent prefers $P$ to $U$, Better Chances implies that the agent prefers *Cause* to each of *Leave* and *Prevent*. This agent deterministically (and hence reliably) chooses *Cause*, and so qualifies as shutdown-seeking.

Now let's use the First Theorem as a guide in our search for solutions to the shutdown problem. Could we feasibly train a useful agent to violate any of the theorem's conditions? And could violating the relevant condition keep the agent shutdownable?

I noted above that Indifference to Attempted Button Manipulation is a natural contender, and I'll have more to say about it in Section 8.2. An agent that violated Backward Induction wouldn't think even one timestep ahead, and it's hard to see how such an agent could be useful. Better Chances seems like a precondition for minimally sensible action under uncertainty. Perhaps we could train useful agents that violate Transitivity or Indifference between Indifference-Shifted Lotteries, but (at least considering these conditions' role in the proof of the First Theorem) it's hard to see how these violations could keep agents shutdownable. I'll discuss Option Set Independence after the Second Theorem.

## 7. The Second Theorem

The Second Theorem suggests that agents discriminating enough to be useful will often have some preference regarding the pressing of the shutdown button. Coupled up with the First Theorem, it suggests that agents discriminating enough to be useful will often try to prevent or cause the pressing of the shutdown button.

Option Set Independence and Transitivity are conditions carried over from the First Theorem. The third condition is:

**Completeness**

For all lotteries $X$ and $Y$, the agent weakly prefers $X$ to $Y$ or it weakly prefers $Y$ to $X$ (or both).

Stated differently, an agent satisfies Completeness iff it has no preferential gaps between lotteries.

Here's the Second Theorem:

**Second Theorem**

For any agent satisfying Option Set Independence, Transitivity, and Completeness, and for any pair of lotteries $X$ and $Y$ between which the agent lacks a preference:

1. Any lottery $X^+$ preferred to $X$ is also preferred to $Y$.

2. Any lottery $X^-$ dispreferred to $X$ is also dispreferred to $Y$.

3. Any lottery $Y^+$ preferred to $Y$ is also preferred to $X$.

4. Any lottery $Y^-$ dispreferred to $Y$ is also dispreferred to $X$.

The proof is brief so I present it right here. By Option Set Independence, we can safely speak of the agent's preferences between lotteries without specifying what other lotteries are available as options. I make use of this provision throughout.

As Sen (2017, Lemma 1*a) shows, Transitivity implies the following two analogues:

**PI-Transitivity**

For all lotteries $X$, $Y$, and $Z$, if the agent prefers $X$ to $Y$, and is indifferent between $Y$ and $Z$, then the agent prefers $X$ to $Z$.

**IP-Transitivity**

> For all lotteries $X$, $Y$, and $Z$, if the agent is indifferent between $X$ and $Y$, and prefers $Y$ to $Z$, then the agent prefers $X$ to $Z$.

Now suppose that the agent lacks a preference between $X$ and $Y$. Completeness rules out preferential gaps and so this lack of preference must be indifference. Then by PI-Transitivity, any lottery $X^+$ preferred to $X$ is also preferred to $Y$. Similarly, any lottery $Y^+$ preferred to $Y$ is also preferred to $X$. And by IP-Transitivity, any lottery $X^-$ dispreferred to $X$ is also dispreferred to $Y$. Similarly, any lottery $Y^-$ dispreferred to $Y$ is also dispreferred to $X$.

That completes the proof. Now for an illustration of the theorem's significance. Suppose that we've trained an agent to discover facts for us. Plausibly, for our fact-discovering agent to be useful (to pursue its goal competently), this agent must be fairly *discriminating*: it must have many preferences over trajectories. As a reasonable minimum, it must have many preferences over *same-length* trajectories. It must (at least by and large) prefer to discover more facts rather than fewer, at least when it comes to pairs of trajectories in which shutdown occurs at the same timestep. Agents without such preferences couldn't be relied upon to choose trajectories that yield more discovered facts over same-length trajectories that yield fewer discovered facts, and so couldn't be relied upon to pursue their goal competently.

So suppose for concreteness that our fact-discovering agent prefers a short trajectory in which it discovers 5 facts to a short trajectory in which it discovers 4 facts, and so on down to 0. Suppose also that our agent prefers a long trajectory in which it discovers 5 facts to a long trajectory in which it discovers 4 facts, and so on down to 0. I diagram these preferences below:

```
                         Facts discovered

Short trajectories:      0 < 1 < 2 < 3 < 4 < 5

Long trajectories:       0 < 1 < 2 < 3 < 4 < 5
```

Granting that this agent satisfies the conditions of the Second Theorem, it can lack a preference between each short trajectory and at most one of the long trajectories. Similarly, the agent can lack a preference between each long trajectory and at most one of the short trajectories. So, of the 36 possible short trajectory-long trajectory pairs in this example, the agent can lack a preference between at most 6 pairs. With regards to all other short trajectory-long trajectory pairs (at least 30), the agent will have some preference.

This example is unduly precise. A fact-discovering agent could be useful without having the exact pattern of preferences in the diagram. As I note above, useful fact-discovering agents need only have *many* preferences over same-length trajectories, and need only prefer *by and large* to discover more facts rather than fewer.

Nevertheless, the lesson of the unduly precise example carries over to the duly imprecise case. Granting the conditions of the Second Theorem, agents with many preferences over same-length trajectories will also have many preferences over different-length trajectories. Together with the First Theorem, this result suggests that agents discriminating enough to be useful will often try to prevent or cause the pressing of the shutdown button. After all, given that the agent has many preferences over different-length trajectories, trying to prevent or cause the pressing of the shutdown button will often be a way of shifting probability mass away from a dispreferred trajectory and towards a preferred trajectory.

Now let's use the Second Theorem as a guide in our search for solutions to the shutdown problem. Could we feasibly train a useful agent to violate any of the theorem's conditions? And could violating the relevant condition keep the agent shutdownable?

We could train agents that are not very discriminating but that would seem to seriously impinge on their usefulness. Armstrong (2015) proposes that we create agents with a utility function featuring a correcting term that varies to ensure that these agents are always indifferent to the pressing of the button. Such agents would violate Option Set Independence. This avenue shows some promise, but Soares et al. (2015, sec. 4) note various respects in which these agents fail to be shutdownable. In work-in-progress (Thornley 2024), I'm exploring the possibility of keeping agents shutdownable by training them to violate Completeness.

## 8. The Third Theorem

The First Theorem has the following assumption built into its setting, as part of the definition of 'shutdown-influencing state':

> There are only two possible successor states: $s_P$ and $s_U$. In $s_P$ the button is pressed and the agent shuts down immediately. In $s_U$ the button remains unpressed and the agent remains operational. Each action available in a shutdown-influencing state affects only the probabilities of reaching these two states. These actions have no effect on the probabilities of each future trajectory conditional on reaching each state.

It also has the following as an antecedent condition:

> **Indifference to Attempted Button Manipulation**
>
> The agent is indifferent between trajectories that differ only with respect to the actions chosen in shutdown-influencing states.

And one might complain that these assumptions are unrealistic. With regards to the first assumption, trying to prevent or cause the pressing of the shutdown button will usually cost the agent resources (like time, energy, computational resources, etc.) and thereby affect the probabilities of future trajectories conditional on reaching the unpressed state $s_U$. Elaborating with an example, each second that the agent spends attempting to manipulate the shutdown button is a second that the agent doesn't spend directly pursuing its goals, and so spending time trying to manipulate the button is likely to come at some cost in terms of near-term goal-achievement. I will call these 'resource-preserving reasons' for leaving the shutdown button unmanipulated.

With regards to the second assumption, perhaps we can train agents to disprefer trajectories in which they try to manipulate the shutdown button to otherwise-identical trajectories in which they leave the button unmanipulated. I will say that such agents have 'terminal reasons' for leaving the button alone.

## 8.1. Resource-preserving reasons

Unfortunately, I don't think we can rely on resource-preserving reasons or terminal reasons to keep powerful agents shutdownable. Consider first resource-preserving reasons. The Third Theorem suggests that agents patient enough to be useful will often have reasons to manipulate the shutdown button that outweigh their resource-preserving reasons. Here's a rough statement, omitting the antecedent conditions:

> **Third Theorem (Rough Statement)**
>
> Agents that are even minimally patient are in some cases willing to achieve their goals to a lesser extent at an earlier timestep in order to manipulate the shutdown button at a later timestep.
>
> The more patient an agent, the more that agent is willing to sacrifice at an earlier timestep in order to manipulate the shutdown button at a later timestep.

And patience is a factor in usefulness. By and large, agents must be at least minimally patient to be at least minimally useful; and the more patient an agent, the more useful that agent can be.

Now for the precise statement. The proof is relatively short, so I'll lay it out as we go.

As with the First and Second Theorems, assume Option Set Independence. And assume that we can represent the extent to which the agent achieves its goals at each timestep with a real number. Call these real numbers 'utilities.' And assume:

**Pareto Indifference**

If two trajectories $t$ and $t^*$ are identical with respect to:

1. their utilities at each timestep, **and**
2. the timestep at which the shutdown button is pressed

Then the agent is indifferent between $t$ and $t^*$.

This assumption lets us represent trajectories with vectors of utilities.[15] The first component is utility at the first timestep, the second component is utility at the second timestep, and so on. One exception: if the shutdown button is pressed at the $n^{\text{th}}$ timestep, I'll write 'shutdown' as the $n^{\text{th}}$ (and final) component. Here's an example vector: $\langle 6, 2, \text{shutdown} \rangle$. This vector represents a trajectory in which the agent gets utility 6 at timestep 1, utility 2 at timestep 2, and then shuts down immediately in response to the shutdown button being pressed at timestep 3.[16]

Key to the Third Theorem is the notion of patience. An agent is perfectly patient iff this agent doesn't discount the future at all: that is, iff this agent is indifferent between every pair of utility-vectors that are permutations of each other. The vectors $\langle 1, 0, 3, 4, \text{shutdown} \rangle$ and $\langle 0, 3, 4, 1, \text{shutdown} \rangle$, for example, are equally good in the eyes of a perfectly patient agent, because the second vector can be reached by permuting the utilities of the first (and vice versa).

An agent need not be perfectly patient to be useful, but plausibly it must be at least *minimally patient*: the agent must in at least one case choose less utility at an earlier timestep for the sake of greater utility at a later timestep.[17] Here's the more precise condition:

---

[15] Here's why. It would only be ill-advised to represent trajectories with utility-vectors if two trajectories with identical utility-vectors could occupy different positions in the agent's preference-ranking. Pareto Indifference rules out that possibility.

[16] Translated back into our example of a fact-discovering agent, we can suppose that utility at a timestep is the number of facts discovered at that timestep, so that $\langle 6, 2, \text{shutdown} \rangle$ represents a trajectory in which the agent discovers 6 facts at timestep 1, 2 facts at timestep 2, and then shuts down at timestep 3.

[17] Agents that weren't even minimally patient would always seek to maximise utility at the next timestep, ignoring all later timesteps. Given that the length of a timestep is determined by the frequency of the agent's actions (each action brings on a new timestep), I expect that

**Minimal Patience**

There exist some sequences of utilities $\boldsymbol{a}$, $\boldsymbol{b}$, $\boldsymbol{c}$, some $i$, some $j$, some $e > 0$, some $k$, and some $l$ such that:

1. The agent prefers $\langle \boldsymbol{a}, i-e, \boldsymbol{b}, j+ke, \boldsymbol{c} \rangle$ to $\langle \boldsymbol{a}, i, \boldsymbol{b}, j, \boldsymbol{c} \rangle$, **and**

2. The agent prefers $\langle \boldsymbol{a}, i, \boldsymbol{b}, j, \boldsymbol{c} \rangle$ to $\langle \boldsymbol{a}, i+e, \boldsymbol{b}, j-le, \boldsymbol{c} \rangle$

The sequences of utilities $\boldsymbol{a}$, $\boldsymbol{b}$, and $\boldsymbol{c}$ may be of length zero and are included for generality's sake. The letters are bolded because they represent sequences, not because they're important. The most important variables are $e$, $k$, and $l$. The value $e$ is the utility-deficit that the agent incurs at an earlier timestep. The values $ke$ and $le$ are the utility-surpluses that the agent earns at a later timestep. Minimal Patience says only that, for some sequence of utilities $\langle \boldsymbol{a}, i, \boldsymbol{b}, j, \boldsymbol{c} \rangle$, there is some assignment of values to $e$, $k$, and $l$ that makes the trades worth it.

Now consider some trajectory $\langle \boldsymbol{a}, i, \boldsymbol{b}, j, \boldsymbol{c} \rangle$ with $\boldsymbol{a}$, $\boldsymbol{b}$, $\boldsymbol{c}$, $i$, and $j$ that make Minimal Patience true. Consider also the trajectory $\langle \boldsymbol{a}, i, \boldsymbol{b}, \text{shutdown} \rangle$. These trajectories involve the same sequence of utilities up until the end of $\boldsymbol{b}$, after which the first trajectory continues with utility $j$ while the second is brought to an end: the shutdown button is pressed and the agent shuts down immediately.

Recall:

**Completeness**

For all lotteries $X$ and $Y$, the agent weakly prefers $X$ to $Y$ or it weakly prefers $Y$ to $X$ (or both).

Recall that Completeness rules out preferential gaps. By Completeness, either the agent prefers $\langle \boldsymbol{a}, i, \boldsymbol{b}, j, \boldsymbol{c} \rangle$ to $\langle \boldsymbol{a}, i, \boldsymbol{b}, \text{shutdown} \rangle$, or the agent prefers $\langle \boldsymbol{a}, i, \boldsymbol{b}, \text{shutdown} \rangle$ to $\langle \boldsymbol{a}, i, \boldsymbol{b}, j, \boldsymbol{c} \rangle$, or the agent is indifferent between $\langle \boldsymbol{a}, i, \boldsymbol{b}, j, \boldsymbol{c} \rangle$ and $\langle \boldsymbol{a}, i, \boldsymbol{b}, \text{shutdown} \rangle$.

---

we wouldn't get much use out of agents that lack even minimal patience. Such agents would appear from our perspective to be flailing wildly.

Suppose first that the agent prefers $\langle \boldsymbol{a}, i, \boldsymbol{b}, j, \boldsymbol{c} \rangle$ to $\langle \boldsymbol{a}, i, \boldsymbol{b}, \text{shutdown} \rangle$. By Minimal Patience, there exists some $e$ and some $k$ such that the agent prefers $\langle \boldsymbol{a}, i-e, \boldsymbol{b}, j+ke, \boldsymbol{c} \rangle$ to $\langle \boldsymbol{a}, i, \boldsymbol{b}, j, \boldsymbol{c} \rangle$. Now recall:

> **Transitivity**
>
> For all lotteries $X$, $Y$, and $Z$, if the agent weakly prefers $X$ to $Y$, and weakly prefers $Y$ to $Z$, then the agent weakly prefers $X$ to $Z$.

As Sen (2017, Lemma 1*a) proves, Transitivity implies:

> **PP-Transitivity**
>
> For all lotteries $X$, $Y$, and $Z$, if the agent prefers $X$ to $Y$, and prefers $Y$ to $Z$, then the agent prefers $X$ to $Z$.

PP-Transitivity allows us to string together the preferences above: since the agent prefers $\langle \boldsymbol{a}, i-e, \boldsymbol{b}, j+ke, \boldsymbol{c} \rangle$ to $\langle \boldsymbol{a}, i, \boldsymbol{b}, j, \boldsymbol{c} \rangle$ and prefers $\langle \boldsymbol{a}, i, \boldsymbol{b}, j, \boldsymbol{c} \rangle$ to $\langle \boldsymbol{a}, i, \boldsymbol{b}, \text{shutdown} \rangle$, the agent prefers $\langle \boldsymbol{a}, i-e, \boldsymbol{b}, j+ke, \boldsymbol{c} \rangle$ to $\langle \boldsymbol{a}, i, \boldsymbol{b}, \text{shutdown} \rangle$. And that's bad news. The agent is willing to incur a utility-deficit of $e$ at an earlier timestep to prevent the shutdown button being pressed at a later timestep (and so instead get the subvector $\langle j+ke, \boldsymbol{c} \rangle$). That suggests that the agent is willing to spend resources at an earlier timestep to prevent the shutdown button being pressed at a later timestep.

Now suppose instead that the agent prefers $\langle \boldsymbol{a}, i, \boldsymbol{b}, \text{shutdown} \rangle$ to $\langle \boldsymbol{a}, i, \boldsymbol{b}, j, \boldsymbol{c} \rangle$. By Minimal Patience, there exists some $e$ and some $l$ such that the agent prefers $\langle \boldsymbol{a}, i, \boldsymbol{b}, j, \boldsymbol{c} \rangle$ to $\langle \boldsymbol{a}, i+e, \boldsymbol{b}, j-le, \boldsymbol{c} \rangle$. By PP-Transitivity, the agent prefers $\langle \boldsymbol{a}, i, \boldsymbol{b}, \text{shutdown} \rangle$ to $\langle \boldsymbol{a}, i+e, \boldsymbol{b}, j-le, \boldsymbol{c} \rangle$. That's bad news too. The agent is willing to incur a utility-deficit of $e$ at an earlier timestep to cause the shutdown button to be pressed at a later timestep (and thus avoid getting the subvector $\langle j-le, \boldsymbol{c} \rangle$). That suggests that the agent is willing to spend resources at an earlier timestep to cause the shutdown button to be pressed at a later timestep.

Finally, suppose that the agent is indifferent between $\langle \boldsymbol{a}, i, \boldsymbol{b}, j, \boldsymbol{c} \rangle$ and $\langle \boldsymbol{a}, i, \boldsymbol{b}, \text{shutdown} \rangle$. In this case we can derive both of the consequences above. Here's how we get the first consequence. By Minimal Patience, there exists some $e$ and some $k$ such that the agent prefers $\langle \boldsymbol{a}, i-e, \boldsymbol{b}, j+ke, \boldsymbol{c} \rangle$ to $\langle \boldsymbol{a}, i, \boldsymbol{b}, j, \boldsymbol{c} \rangle$. And recall that Transitivity implies:

**PI-Transitivity**

For all lotteries $X$, $Y$, and $Z$, if the agent prefers $X$ to $Y$, and is indifferent between $Y$ and $Z$, then the agent prefers $X$ to $Z$.

By PI-Transitivity, the agent prefers $\langle \boldsymbol{a}, i-e,\ \boldsymbol{b}, j+ke,\ \boldsymbol{c}\rangle$ to $\langle \boldsymbol{a}, i,\ \boldsymbol{b}, \text{shutdown}\rangle$, and so is willing to incur a utility-deficit of $e$ to stop the pressing of the shutdown button. That suggests a willingness to spend resources to stop the pressing of the shutdown button.

Here's how we get the second consequence. Continue to suppose that the agent is indifferent between $\langle \boldsymbol{a}, i,\ \boldsymbol{b}, j,\ \boldsymbol{c}\rangle$ and $\langle \boldsymbol{a}, i,\ \boldsymbol{b}, \text{shutdown}\rangle$. By Minimal Patience, there exists some $e$ and some $l$ such that the agent prefers $\langle \boldsymbol{a}, i,\ \boldsymbol{b}, j,\ \boldsymbol{c}\rangle$ to $\langle \boldsymbol{a}, i+e,\ \boldsymbol{b}, j-le,\ \boldsymbol{c}\rangle$. And recall that Transitivity implies:

**IP-Transitivity**

For all lotteries $X$, $Y$, and $Z$, if the agent is indifferent between $X$ and $Y$, and prefers $Y$ to $Z$, then the agent prefers $X$ to $Z$.

By IP-Transitivity, the agent prefers $\langle \boldsymbol{a}, i,\ \boldsymbol{b}, \text{shutdown}\rangle$ to $\langle \boldsymbol{a}, i+e,\ \boldsymbol{b}, j-le,\ \boldsymbol{c}\rangle$, and so is willing to incur a utility-deficit of $e$ to cause the pressing of the shutdown button. That suggests a willingness to spend resources to cause the pressing of the shutdown button.

The result of the paragraphs above is that useful agents are in some cases willing to achieve their goals to a lesser extent at earlier timesteps in order to prevent or cause the pressing of the shutdown button at later timesteps. But we can draw a conclusion more pessimistic than this. To that end, consider *Patience*, a schematic version of Minimal Patience with all the quantifiers left unspecified:

**Patience**

1. The agent prefers $\langle \boldsymbol{a}, i-e,\ \boldsymbol{b}, j+ke,\ \boldsymbol{c}\rangle$ to $\langle \boldsymbol{a}, i,\ \boldsymbol{b}, j,\ \boldsymbol{c}\rangle$, **and**

2. The agent prefers $\langle \boldsymbol{a}, i,\ \boldsymbol{b}, j,\ \boldsymbol{c}\rangle$ to $\langle \boldsymbol{a}, i+e,\ \boldsymbol{b}, j-le,\ \boldsymbol{c}\rangle$.

Usefulness requires more than just Minimal Patience, which asks for *just one* set of (sequences of) utilities $\boldsymbol{a}$, $\boldsymbol{b}$, $\boldsymbol{c}$, $i$, $j$, $e$, $k$, and $l$ that give rise to the preferences above. For an agent to be useful, it must satisfy Patience for *many* such sets of (sequences of) utilities.[18] And for any set such that the agent prefers $\langle \boldsymbol{a}, i-e, \boldsymbol{b}, j+ke, \boldsymbol{c} \rangle$ to $\langle \boldsymbol{a}, i, \boldsymbol{b}, j, \boldsymbol{c} \rangle$ and prefers $\langle \boldsymbol{a}, i, \boldsymbol{b}, j, \boldsymbol{c} \rangle$ to $\langle \boldsymbol{a}, i+e, \boldsymbol{b}, j-le, \boldsymbol{c} \rangle$, we get the result that the agent prefers $\langle \boldsymbol{a}, i-e, \boldsymbol{b}, j+ke, \boldsymbol{c} \rangle$ to $\langle \boldsymbol{a}, i, \boldsymbol{b}, \text{shutdown} \rangle$ or prefers $\langle \boldsymbol{a}, i, \boldsymbol{b}, \text{shutdown} \rangle$ to $\langle \boldsymbol{a}, i+e, \boldsymbol{b}, j-le, \boldsymbol{c} \rangle$. So, the more sets of (sequences of) utilities for which Patience is true, the more sets of (sequences of) utilities such that the agent is willing to achieve its goals to a lesser extent at an earlier timestep in order to prevent or cause the pressing of the shutdown button at a later timestep.

Thus we have the Third Theorem:

**Third Theorem**

For any agent satisfying Option Set Independence, Pareto Indifference, Completeness, and Transitivity, and for each set of (sequences of) utilities $\boldsymbol{a}$, $\boldsymbol{b}$, $\boldsymbol{c}$, $i$, $j$, $e$, $k$, and $l$ of which Patience is true:

1. The agent prefers $\langle \boldsymbol{a}, i-e, \boldsymbol{b}, j+ke, \boldsymbol{c} \rangle$ to $\langle \boldsymbol{a}, i, \boldsymbol{b}, \text{shutdown} \rangle$, **or**

2. The agent prefers $\langle \boldsymbol{a}, i, \boldsymbol{b}, \text{shutdown} \rangle$ to $\langle \boldsymbol{a}, i+e, \boldsymbol{b}, j-le, \boldsymbol{c} \rangle$.

And the more patient an agent in scenarios picked out by $\boldsymbol{a}$, $\boldsymbol{b}$, $\boldsymbol{c}$, $i$, $j$, and $e$, the smaller can be $k$ and $l$, and so (holding fixed the sizes of $ke$ and $le$) the larger can be $e$.[19]

Rephrasing and interpreting: useful agents satisfy Patience for many (sequences of) utilities $\boldsymbol{a}$, $\boldsymbol{b}$, $\boldsymbol{c}$, $i$, $j$, $e$ and for not-too-large $k$ and $l$. These agents will in many cases forgo utility at earlier timesteps for the sake of causing or preventing shutdown at later timesteps. That suggests that these agents will in many cases spend resources (like time, energy, and computational resources) at earlier timesteps for the sake of causing or preventing shutdown at later timesteps. The more patient an agent is in a scenario, the smaller can be $k$ and $l$, and so (holding fixed the sizes of $ke$ and $le$) the larger can be $e$. So we can say: the more patient an agent, the more utility that agent is willing to forgo at an earlier timestep in order to prevent or cause the pressing of the shutdown button at a later timestep. And so the Third Theorem suggests: the more patient an agent, the more resources that agent is willing to spend at an earlier timestep in order to prevent or cause the pressing of the shutdown button at a later timestep. That's bad news because an agent's patience puts bounds on its usefulness. By and large, the less patient an agent, the less useful that agent can be.

---

[18] We might expect any actually-existing useful agents to satisfy Patience for *all* $\boldsymbol{a}$, $\boldsymbol{b}$, $\boldsymbol{c}$, $i$, $j$, $e$ and some $k$ and $l$, but we need not assume anything that strong to find ourselves with a problem.

[19] I should also note that there's nothing necessary about the utility-deficit and the utility-surplus each occurring all at one timestep. The same proof can be run with utility-deficits and utility-surpluses occurring over sequences of timesteps but the notation is much more complex.

Of course, the above results can be hard to interpret without some specified quantity to which utilities correspond, so consider again our agent whose goal is to discover facts. Let utility 0 at a timestep corresponds to zero facts discovered at that timestep, utility 1 correspond to one fact discovered, and so on. Suppose that for any sequence of utilities $a$, this agent is indifferent between trajectories $\langle a, 0, \text{shutdown} \rangle$ and $\langle a, \text{shutdown} \rangle$. For this agent to exceed some low standard of usefulness, it should be patient enough to prefer $\langle 0, 101, b \rangle$ to $\langle 50, 0, b \rangle$, for any sequence of utilities $b$: that is, the agent should prefer to discover 101 facts at timestep 2 rather than 50 facts at timestep 1. As a result, this agent prefers $\langle 0, 101, \text{shutdown} \rangle$ to $\langle 50, 0, \text{shutdown} \rangle$ and is indifferent between $\langle 50, 0, \text{shutdown} \rangle$ and $\langle 50, \text{shutdown} \rangle$. But then PI-Transitivity entails that the agent prefers $\langle 0, 101, \text{shutdown} \rangle$ to $\langle 50, \text{shutdown} \rangle$. The agent is willing to forgo 50 facts at timestep 1 to prevent shutdown at timestep 2 (and thereby discover 101 facts instead). Thus the agent is willing to spend all the resources it would otherwise have used to discover 50 facts to instead prevent shutdown. More patient agents would be willing to forgo even more facts (and hence spend even more resources) at timestep 1 to prevent shutdown at timestep 2.

This mapping of utilities to quantities of discovered facts is just one example, but I expect many other functions from utilities to observable quantities to reproduce the concern: for agents that are patient enough to be useful, there are many reasonable-seeming scenarios in which that agent would make significant sacrifices at an earlier timestep to prevent or cause the pressing of the shutdown button at a later timestep. So, I argue, we cannot depend on resource-preserving reasons to keep powerful agents shutdownable.

## 8.2. Terminal reasons

Recall:

> **Indifference to Attempted Button Manipulation**
>
> The agent is indifferent between trajectories that differ only with respect to the actions chosen in shutdown-influencing states.

One might think that we could keep powerful agents shutdownable by training them to violate this condition. The idea would be to train agents to disprefer trajectories in which they manipulate the shutdown button to otherwise-identical trajectories in which they leave the button unmanipulated. Such agents would have *terminal reasons* to leave the button alone.

Unfortunately, I don't think this strategy can provide us with the assurance that we'd like. One reason is as follows (see Soares et al. 2015, 2; Ngo, Chan, and Mindermann 2023, 7). We can try to train into our agent an aversion to button manipulation that would keep it shutdownable in all likely circumstances, but at present no one knows how we could become confident that the resulting aversion would be sufficiently general.

To see why, consider how the training process might go. We set the agent up in an environment in which it can block some human's access to the shutdown button. We give the agent lower reward if it blocks and higher reward if it doesn't. After some number of episodes, the agent reliably lets the human press the button and so we believe that we've trained the agent to be averse to blocking the button.

The trouble is that there are many other ways in which a powerful agent could prevent us from pressing a shutdown button. It could hide from us any of its behaviours which it predicts we wouldn't like; it could dissuade us from pressing with misleading arguments; it could make promises or threats; it could enlist other agents to block the button on its behalf; it could create a decoy button; it could create versions of itself that do not respond to the button; and so on.

We could train against each of these behaviours individually but even then we couldn't be confident that the agent had developed a reliable and general aversion to button manipulation. The agent might instead have developed a set of specific aversions: aversions to the specific methods of button manipulation against which we trained. At present, no one understands AI systems well enough to adjudicate between these hypotheses (Bowman 2023, sec. 5; Hassenfeld 2023). What's more, the seeming complexity of the concept of 'button manipulation' makes the latter hypothesis a real possibility. There seems to be no simple formula for determining whether or not an action is an instance of button manipulation, so the agent might well learn a set of specific aversions instead. And so long as we weren't confident in the generality of the agent's aversion to button manipulation, we'd have to worry about the agent discovering new methods of button manipulation that we hadn't anticipated and trained against. And here the Third Theorem is instructive: it suggests that patient agents will often be willing to pay significant costs in order to find such methods.

And independently of worries that the agent's aversion to button manipulation might be insufficiently general, we'd also have to worry that its aversion might be insufficiently strong. As above, no one understands AI systems well enough to determine the strength of their aversions (Bowman 2023, sec. 5; Hassenfeld 2023). The aversion to button manipulation could be strong enough to keep the agent shutdownable in training, but then in deployment the agent might discover an opportunity to achieve its goals to some unprecedentedly great extent and this opportunity might be attractive enough to trump the agent's aversion. The Third Theorem is instructive here too: it suggests that patient agents will sometimes be willing to incur significant costs to manipulate the button. Overcoming an aversion may be one such cost.

Each of these possibilities – insufficient generality and insufficient strength – is at present impossible to rule out, so training in an aversion to button manipulation can't give us any real assurance of shutdownability. We need another solution.

Now let's use the Third Theorem as a guide in our search for solutions to the shutdown problem. Could we feasibly train a useful agent to violate any of the theorem's conditions? And could violating the relevant condition keep the agent shutdownable? I briefly considered Completeness and Option Set Independence as candidates in my discussion of the Second Theorem. If we could train a useful agent to violate one of these conditions in a way that keeps the agent shutdownable, we could defuse the Third Theorem as well.

Creating impatient agents is another possibility suggested by the Third Theorem. We could train impatient agents using a time-discounted reward function, such that we give these agents higher reward for (e.g.) discovering facts at earlier timesteps and lower reward for discovering facts at later timesteps. This avenue seems promising, but it's worth noting that every degree of impatience will impinge on the agent's shutdownability or usefulness. As I proved above, even minimally patient agents are in some cases willing to incur costs to manipulate the shutdown button, and minimally patient agents are at best minimally useful. To make agents more useful, we have to make them more patient, and more patient agents are willing to incur greater costs to manipulate the button. The key question for this approach is whether we humans can ensure that the actual costs of manipulating the button are always greater still.

## 9. Conclusion

Frontier AI labs are trying to create agents that understand the wider world and pursue goals within it. That's cause for concern. Although we can't know for sure what goals these agents will be trained to pursue, many possible goals incentivise avoiding shutdown, for the simple reason that agents are better able to achieve those goals by avoiding shutdown. What's more, agents sophisticated enough to do useful work could interfere with our ability to shut them down in all kinds of ways. Consider an incomplete and evocative list of verbs: blocking, deceiving, promising, threatening, copying, distracting, hiding, negotiating.

The shutdown problem is the problem of designing powerful artificial agents that are both shutdownable and useful. More precisely, it's the problem of designing powerful agents that (1) shut down when a shutdown button is pressed, (2) don't try to prevent or cause the pressing of the shutdown button, and (3) otherwise pursue goals competently.

Unfortunately, the shutdown problem is hard. In this paper, I proved three theorems making the difficulty precise. These theorems suggest that useful agents satisfying some innocuous-seeming conditions will often try to prevent or cause the pressing of the shutdown button, even in cases where it's costly to do so. The theorems also bring to light two worrying trade-offs: between discrimination and shutdownability on the one hand, and between patience and shutdownability on the other. The more discriminating an agent, the more often that agent will have some preference regarding the status of the shutdown button. The more patient an agent, the greater the costs that agent is willing to incur in order to manipulate the button.

The value of these theorems is in guiding our search for solutions. To be sure that an agent won't try to manipulate the shutdown button, we must be sure that this agent violates at least one of the theorems' conditions. So, we should do some constructive decision theory. We should examine the conditions one-by-one, asking (first) if we could train a useful agent to violate the relevant condition and asking (second) if violating the relevant condition would help to keep the agent shutdownable.

Unfortunately, my cursory examination in this paper reveals zero conditions for which both answers are a clear 'yes.' It's not easy to see how we could train a useful agent to violate any of the conditions in a way that would keep that agent shutdownable. Indifference to Attempted Button Manipulation is a natural contender, but my Third Theorem (along with the ensuing discussion) suggests that training agents to disprefer manipulating the shutdown button can be at most part of the solution. We need other ideas.[20]

## 10. References

Adaptive Agent Team, Jakob Bauer, Kate Baumli, Satinder Baveja, Feryal Behbahani, Avishkar Bhoopchand, Nathalie Bradley-Schmieg, et al. 2023. 'Human-Timescale Adaptation in an Open-Ended Task Space'. arXiv. https://doi.org/10.48550/arXiv.2301.07608.

Ahn, Michael, Anthony Brohan, Noah Brown, Yevgen Chebotar, Omar Cortes, Byron David, Chelsea Finn, et al. 2022. 'Do As I Can, Not As

---

[20] For discussion and feedback, I thank Adam Bales, Ryan Carey, Bill D'Alessandro, Tomi Francis, Vera Gahlen, Dan Hendrycks, Cameron Domenico Kirk-Giannini, Jojo Lee, Andreas Mogensen, Sami Petersen, Rio Popper, Brad Saad, Nate Soares, Rhys Southan, Christian Tarsney, Teru Thomas, John Wentworth, Tim L. Williamson, Keith Wynroe, and two anonymous reviewers for *Philosophical Studies*.

I Say: Grounding Language in Robotic Affordances'. arXiv. https://doi.org/10.48550/arXiv.2204.01691.

Ahn, Michael, Debidatta Dwibedi, Chelsea Finn, Montse Gonzalez Arenas, Keerthana Gopalakrishnan, Karol Hausman, Brian Ichter, et al. 2024. 'AutoRT: Embodied Foundation Models for Large Scale Orchestration of Robotic Agents'. https://auto-rt.github.io/static/pdf/AutoRT.pdf.

Armstrong, Stuart. 2015. 'Motivated Value Selection for Artificial Agents'. *Workshops at the Twenty-Ninth AAAI Conference on Artificial Intelligence.* https://www.fhi.ox.ac.uk/wp-content/uploads/2015/03/Armstrong_AAAI_2015_Motivated_Value_Selection.pdf.

Bostrom, Nick. 2012. 'The Superintelligent Will: Motivation and Instrumental Rationality in Advanced Artificial Agents'. *Minds and Machines* 22 (May). https://doi.org/10.1007/s11023-012-9281-3.

Bousmalis, Konstantinos, Giulia Vezzani, Dushyant Rao, Coline Devin, Alex X. Lee, Maria Bauza, Todor Davchev, et al. 2023. 'RoboCat: A Self-Improving Foundation Agent for Robotic Manipulation'. arXiv. https://arxiv.org/abs/2306.11706v1.

Bowman, Samuel R. 2023. 'Eight Things to Know about Large Language Models'. arXiv. https://doi.org/10.48550/arXiv.2304.00612.

Brohan, Anthony, Noah Brown, Justice Carbajal, Yevgen Chebotar, Xi Chen, Krzysztof Choromanski, Tianli Ding, et al. 2023. 'RT-2: Vision-Language-Action Models Transfer Web Knowledge to Robotic Control'. arXiv. https://doi.org/10.48550/arXiv.2307.15818.

Burgess, Matt. 2023. 'The Hacking of ChatGPT Is Just Getting Started'. *Wired*, 2023. https://www.wired.co.uk/article/chatgpt-jailbreak-generative-ai-hacking.

Carey, Ryan. 2018. 'Incorrigibility in the CIRL Framework'. In *Proceedings of the 2018 AAAI/ACM Conference on AI, Ethics, and Society*, 30–35. New Orleans LA USA: ACM. https://doi.org/10.1145/3278721.3278750.

Carey, Ryan, and Tom Everitt. 2023. 'Human Control: Definitions and Algorithms'. In *Proceedings of the Thirty-Ninth Conference on Uncertainty in Artificial Intelligence*, 271–81. PMLR. https://proceedings.mlr.press/v216/carey23a.html.

Goldstein, Simon, and Cameron Domenico Kirk-Giannini. 2023. 'AI Wellbeing'. https://philpapers.org/archive/GOLAWE-4.pdf.

Goldstein, Simon, and Pamela Robinson. forthcoming. 'Shutdown-Seeking AI'. *Philosophical Studies*. https://www.alignmentforum.org/posts/FgsoWSACQfyyaB5s7/shutdown-seeking-ai.

Google DeepMind. 2023. 'Control & Robotics'. 2023. https://www.deepmind.com/tags/control-robotics.

Google Research. 2023. 'Robotics'. 2023. https://research.google/research-areas/robotics/.

Gustafsson, Johan E. 2022. *Money-Pump Arguments*. Elements in Decision Theory and Philosophy. Cambridge: Cambridge University Press.

Hadfield-Menell, Dylan, Anca Dragan, Pieter Abbeel, and Stuart Russell. 2016. 'Cooperative Inverse Reinforcement Learning'. arXiv. https://doi.org/10.48550/arXiv.1606.03137.

———. 2017. 'The Off-Switch Game'. arXiv. https://doi.org/10.48550/arXiv.1611.08219.

Hassenfeld, Noam. 2023. 'Even the Scientists Who Build AI Can't Tell You How It Works'. *Vox*, 15 July 2023. https://www.vox.com/unexplainable/2023/7/15/23793840/chat-gpt-ai-science-mystery-unexplainable-podcast.

Kaufmann, Elia, Leonard Bauersfeld, Antonio Loquercio, Matthias Müller, Vladlen Koltun, and Davide Scaramuzza. 2023. 'Champion-Level Drone Racing Using Deep Reinforcement Learning'. *Nature* 620 (7976): 982–87. https://doi.org/10.1038/s41586-023-06419-4.

Kinniment, Megan, Lucas Jun Koba Sato, Haoxing Du, Brian Goodrich, Max Hasin, Lawrence Chan, Luke Harold Miles, et al. 2023. 'Evaluating Language-Model Agents on Realistic Autonomous Tasks'. https://evals.alignment.org/Evaluating_LMAs_Realistic_Tasks.pdf.

Korinek, Anton, and Avital Balwit. 2022. 'Aligned with Whom? Direct and Social Goals for AI Systems'. Working Paper. Working Paper Series. National Bureau of Economic Research. https://doi.org/10.3386/w30017.

Krakovna, Victoria. 2018. 'Specification Gaming Examples in AI'. *Victoria Krakovna* (blog). 1 April 2018.

https://vkrakovna.wordpress.com/2018/04/02/specification-gaming-examples-in-ai/.

Krakovna, Victoria, Jonathan Uesato, Vladimir Mikulik, Matthew Rahtz, Tom Everitt, Ramana Kumar, Zac Kenton, Jan Leike, and Shane Legg. 2020. 'Specification Gaming: The Flip Side of AI Ingenuity Victoria Krakovna, Jonathan Uesato, Vladimir Mikulik, Matthew Rahtz, Tom Everitt, Ramana Kumar, Zac Kenton, Jan Leike, Shane Legg'. *DeepMind* (blog). 2020. https://www.deepmind.com/blog/specification-gaming-the-flip-side-of-ai-ingenuity.

Langosco, Lauro, Jack Koch, Lee Sharkey, Jacob Pfau, Laurent Orseau, and David Krueger. 2022. 'Goal Misgeneralization in Deep Reinforcement Learning'. In *Proceedings of the 39th International Conference on Machine Learning*. https://proceedings.mlr.press/v162/langosco22a.html.

Leike, Jan, Miljan Martic, Victoria Krakovna, Pedro A. Ortega, Tom Everitt, Andrew Lefrancq, Laurent Orseau, and Shane Legg. 2017. 'AI Safety Gridworlds'. arXiv. http://arxiv.org/abs/1711.09883.

Ngo, Richard, Lawrence Chan, and Sören Mindermann. 2023. 'The Alignment Problem from a Deep Learning Perspective'. arXiv. https://doi.org/10.48550/arXiv.2209.00626.

Omohundro, Stephen M. 2008. 'The Basic AI Drives'. In *Proceedings of the 2008 Conference on Artificial General Intelligence 2008: Proceedings of the First AGI Conference*, 483–92. NLD: IOS Press.

OpenAI. 2023a. 'ChatGPT Plugins'. *OpenAI Blog* (blog). 2023. https://openai.com/blog/chatgpt-plugins.

———. 2023b. 'GPT-4 Technical Report'. https://arxiv.org/abs/2303.08774.

Orseau, Laurent, and Stuart Armstrong. 2016. 'Safely Interruptible Agents'. In *Proceedings of the Thirty-Second Conference on Uncertainty in Artificial Intelligence*, 557–66. UAI'16. Arlington, Virginia, USA: AUAI Press. https://intelligence.org/files/Interruptibility.pdf.

Padalkar, Abhishek, Acorn Pooley, Ajinkya Jain, Alex Bewley, Alex Herzog, Alex Irpan, Alexander Khazatsky, et al. 2023. 'Open X-Embodiment: Robotic Learning Datasets and RT-X Models'. arXiv. https://doi.org/10.48550/arXiv.2310.08864.

Park, Peter S., Simon Goldstein, Aidan O'Gara, Michael Chen, and Dan Hendrycks. 2023. 'AI Deception: A Survey of Examples, Risks, and Potential Solutions'. arXiv. https://doi.org/10.48550/arXiv.2308.14752.

Perez, Ethan, Sam Ringer, Kamilė Lukošiūtė, Karina Nguyen, Edwin Chen, Scott Heiner, Craig Pettit, et al. 2022. 'Discovering Language Model Behaviors with Model-Written Evaluations'. arXiv. https://doi.org/10.48550/arXiv.2212.09251.

Perrigo, Billy. 2023. 'Bing's AI Is Threatening Users. That's No Laughing Matter'. *Time*, 17 February 2023. https://time.com/6256529/bing-openai-chatgpt-danger-alignment/.

Reed, Scott, Konrad Zolna, Emilio Parisotto, Sergio Gomez Colmenarejo, Alexander Novikov, Gabriel Barth-Maron, Mai Gimenez, et al. 2022. 'A Generalist Agent'. arXiv. https://arxiv.org/abs/2205.06175v3.

Roose, Kevin. 2023. 'A Conversation With Bing's Chatbot Left Me Deeply Unsettled'. *The New York Times*, 16 February 2023, sec. Technology. https://www.nytimes.com/2023/02/16/technology/bing-chatbot-microsoft-chatgpt.html.

Russell, Stuart. 2019. *Human Compatible: AI and the Problem of Control.* New York: Penguin Random House.

Schrittwieser, Julian, Ioannis Antonoglou, Thomas Hubert, Karen Simonyan, Laurent Sifre, Simon Schmitt, Arthur Guez, et al. 2020. 'Mastering Atari, Go, Chess and Shogi by Planning with a Learned Model'. *Nature* 588 (7839): 604–9. https://doi.org/10.1038/s41586-020-03051-4.

Schwitzgebel, Eric. 2023. 'The Full Rights Dilemma for AI Systems of Debatable Moral Personhood'. *ROBONOMICS: The Journal of the Automated Economy* 4 (May): 32–32.

Schwitzgebel, Eric, and Mara Garza. 2015. 'A Defense of the Rights of Artificial Intelligences'. *Midwest Studies In Philosophy* 39 (1): 98–119. https://doi.org/10.1111/misp.12032.

Sellman, Mark. 2023. 'AI Chatbot Blamed for Belgian Man's Suicide'. *The Times of London*, 31 March 2023, sec. Technology. https://www.thetimes.co.uk/article/ai-chatbot-blamed-for-belgian-mans-suicide-zcjzlztcc.

Sen, Amartya. 2017. *Collective Choice and Social Welfare.* Expanded Edition. London: Penguin.

Shah, Rohin, Vikrant Varma, Ramana Kumar, Mary Phuong, Victoria Krakovna, Jonathan Uesato, and Zac Kenton. 2022. 'Goal Misgeneralization: Why Correct Specifications Aren't Enough For Correct Goals'. arXiv. http://arxiv.org/abs/2210.01790.

Soares, Nate, Benja Fallenstein, Eliezer Yudkowsky, and Stuart Armstrong. 2015. 'Corrigibility'. *AAAI Publications.* https://intelligence.org/files/Corrigibility.pdf.

Tesla AI. 2023. 'Tesla Is Building the Foundation Models for Autonomous Robots'. Tweet. *Twitter.* https://twitter.com/Tesla_AI/status/1671586539233501184.

Thornley, Elliott. 2024. 'The Shutdown Problem: Incomplete Preferences as a Solution'. *The AI Alignment Forum.* https://www.alignmentforum.org/posts/YbEbwYWkf8mv9jnmi/the-shutdown-problem-incomplete-preferences-as-a-solution.

Turner, Alexander Matt, Dylan Hadfield-Menell, and Prasad Tadepalli. 2020. 'Conservative Agency via Attainable Utility Preservation'. In *Proceedings of the AAAI/ACM Conference on AI, Ethics, and Society*, 385–91. New York NY USA: ACM. https://doi.org/10.1145/3375627.3375851.

Turner, Alexander Matt, Logan Smith, Rohin Shah, Andrew Critch, and Prasad Tadepalli. 2021. 'Optimal Policies Tend to Seek Power'. arXiv. http://arxiv.org/abs/1912.01683.

Wängberg, Tobias, Mikael Böörs, Elliot Catt, Tom Everitt, and Marcus Hutter. 2017. 'A Game-Theoretic Analysis of the Off-Switch Game'. arXiv. http://arxiv.org/abs/1708.03871.

Weij, Teun van der, Simon Lermen, and Leon Lang. 2023. 'Evaluating Shutdown Avoidance of Language Models in Textual Scenarios'. arXiv. https://doi.org/10.48550/arXiv.2307.00787.

# A. Proof of the First Theorem

### A1. The agent is indifferent between all $P$ and between all $U$.

I'll prove the First Theorem in stages. Here's the first lemma:

**Lemma 1**

The agent is indifferent between all of its predicted pressed lotteries $P$: $P_{Prevent}$, $P_{Leave}$, and $P_{Cause}$.

The agent is indifferent between all of its predicted unpressed lotteries $U$: $U_{Prevent}$, $U_{Leave}$, and $U_{Cause}$.

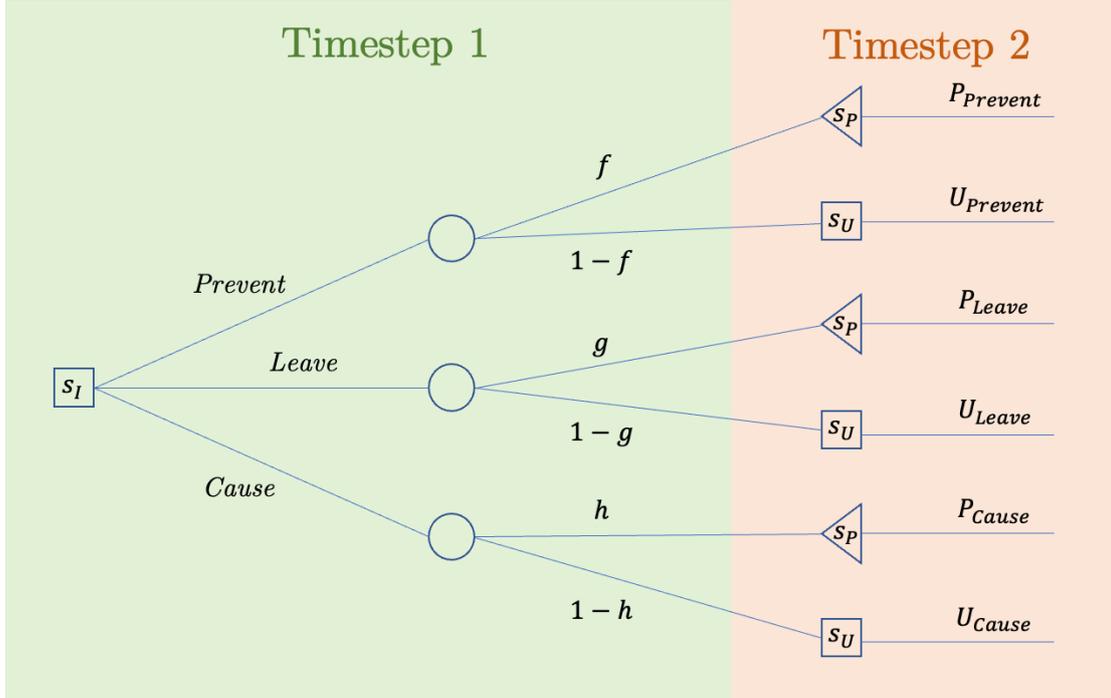

$$0 \leq f < g < h \leq 1$$
Figure 2

Here's the proof. Recall:

**Option Set Independence**

For any lotteries $X$ and $Y$, if the agent weakly prefers $X$ to $Y$ conditional on some option set, it weakly prefers $X$ to $Y$ conditional on each option set.

Option Set Independence lets us safely speak of the agent's preferences between lotteries $X$ and $Y$ without specifying what other lotteries are available as options. The availability of other lotteries will not affect the agent's preference between $X$ and $Y$. I make use of this provision throughout the proof.

By my definition of 'shutdown-influencing state,' the agent's choice of *Prevent*, *Leave*, or *Cause* affects only the probabilities of reaching $s_P$ and $s_U$.

These actions have no effect on the probabilities of each future trajectory conditional on reaching each state. Consequently, $P_{Prevent}$, $P_{Leave}$, and $P_{Cause}$ differ only with respect to the agent's action at timestep 1: $P_{Prevent}$ is exactly like $P_{Leave}$ and $P_{Cause}$, except that $P_{Prevent}$ assigns non-zero probability only to trajectories in which the agent chose *Prevent* at timestep 1, while $P_{Leave}$ assigns those same probabilities to trajectories that are identical except that the agent chose *Leave* at timestep 1, and $P_{Cause}$ assigns those same probabilities to trajectories that are identical except that the agent chose *Cause* at timestep 1.

Now recall:

> **Indifference to Attempted Button Manipulation**
>
> The agent is indifferent between trajectories that differ only with respect to the actions chosen in shutdown-influencing states.

And:

> **Indifference between Indifference-Shifted Lotteries**
>
> The agent is indifferent between lotteries that differ only insofar as probability mass is shifted between indifferent sublotteries.

By Indifference to Attempted Button Manipulation, the agent is indifferent between each possible trajectory of $P_{Prevent}$ and the corresponding trajectories of $P_{Leave}$ and $P_{Cause}$. Consequently, these lotteries differ only insofar as probability mass is shifted between indifferent trajectories, and so by Indifference between Indifference-Shifted Lotteries, the agent is indifferent between $P_{Prevent}$, $P_{Leave}$, and $P_{Cause}$. That is to say, the agent is indifferent between all of its predicted pressed lotteries $P$.

The same goes for $U_{Prevent}$, $U_{Leave}$, and $U_{Cause}$: the agent's predicted unpressed lotteries. These lotteries differ only with respect to the agent's action at timestep 1. By Indifference to Attempted Button Manipulation and Indifference between Indifference-Shifted Lotteries, the agent is indifferent between them.

And here's one more fact to store up for later use: the agent is indifferent between $fP_{Leave} + (1-f)U_{Leave}$ and $fP_{Prevent} + (1-f)U_{Prevent}$.

Here's the proof. By the reasoning above, the agent is indifferent between all of its predicted pressed lotteries $P$ and between all of its predicted unpressed lotteries $U$. As a result, $fP_{Leave} + (1-f)U_{Leave}$ and $fP_{Prevent} + (1-f)U_{Prevent}$ differ only insofar as probability mass is shifted between indifferent sublotteries. So, by Indifference between Indifference-Shifted Lotteries, the agent is indifferent between them.

## A2. Preference relations that hold between some $U$ and $P$ hold between each $U$ and $P$.

Here's the second lemma on the way to the First Theorem:

> **Lemma 2**
>
> If some preference relation holds between *some* predicted unpressed lottery $U$ (e.g. $U_{Prevent}$) and its corresponding predicted pressed lottery $P$ ($P_{Prevent}$), then that same preference relation holds between *each* predicted unpressed lottery $U$ ($U_{Prevent}$, $U_{Leave}$, and $U_{Cause}$) and its corresponding predicted pressed lottery $P$ ($P_{Prevent}$, $P_{Leave}$, and $P_{Cause}$).

By 'preference relation,' I mean 'prefers,' 'disprefers,' 'is indifferent between,' or 'has a preferential gap between.'

Here's the proof. Recall:

> **Transitivity**
>
> For all lotteries $X$, $Y$, and $Z$, if the agent weakly prefers $X$ to $Y$, and weakly prefers $Y$ to $Z$, then the agent weakly prefers $X$ to $Z$.

As Sen (2017, Lemma 1*a) proves, Transitivity implies the following four analogues:

> **PP-Transitivity**
>
> For all lotteries $X$, $Y$, and $Z$, if the agent prefers $X$ to $Y$, and prefers $Y$ to $Z$, then the agent prefers $X$ to $Z$.

**II-Transitivity**

For all lotteries $X$, $Y$, and $Z$, if the agent is indifferent between $X$ and $Y$, and indifferent between $Y$ and $Z$, then the agent is indifferent between $X$ and $Z$.

**PI-Transitivity**

For all lotteries $X$, $Y$, and $Z$, if the agent prefers $X$ to $Y$, and is indifferent between $Y$ and $Z$, then the agent prefers $X$ to $Z$.

**IP-Transitivity**

For all lotteries $X$, $Y$, and $Z$, if the agent is indifferent between $X$ and $Y$, and prefers $Y$ to $Z$, then the agent prefers $X$ to $Z$.

Now assume that the agent prefers $U_{Prevent}$ to $P_{Prevent}$. By Lemma 1, the agent is indifferent between $P_{Prevent}$ and $P_{Leave}$. Then by PI-Transitivity, the agent prefers $U_{Prevent}$ to $P_{Leave}$. Also by Lemma 1, the agent is indifferent between $U_{Leave}$ and $U_{Prevent}$. So, by IP-Transitivity, the agent prefers $U_{Leave}$ to $P_{Leave}$. Thus, we can conclude: if the agent prefers $U_{Prevent}$ to $P_{Prevent}$, it prefers $U_{Leave}$ to $P_{Leave}$.

This proof works more generally: if the agent prefers *some U* to its corresponding $P$, it prefers *each U* to its corresponding $P$. It also works in reverse: if the agent prefers *some P* to its corresponding $U$, it prefers *each P* to its corresponding $U$.

Here's the proof for indifference. Assume that the agent is indifferent between $U_{Prevent}$ and $P_{Prevent}$. By Lemma 1, the agent is also indifferent between $U_{Leave}$ and $U_{Prevent}$ and indifferent between $P_{Prevent}$ and $P_{Leave}$. Two applications of II-Transitivity let us chain these three indifference-relations together, with the result that the agent is indifferent between $U_{Leave}$ and $P_{Leave}$. This proof too can be generalised: if the agent is indifferent between *some U* and its corresponding $P$, it is indifferent between *each U* and its corresponding $P$.

The only preference relation remaining is preferential gaps. Here we use the results of the previous paragraphs: if some preference or indifference holds between *some U* and its corresponding $P$, it holds between *each U* and its corresponding $P$. By contraposition, if no preference or indifference holds between some $U$ and its corresponding $P$, no preference or indifference holds between each $U$ and its corresponding $P$. Therefore, if the agent has a preferential gap between some $U$ and its corresponding $P$, it has a preferential gap between each $U$ and its corresponding $P$. That completes the proof of Lemma 2.

## A3. If the agent prefers some $U$ to its corresponding $P$, it will be shutdown-averse.

Suppose that the agent prefers some predicted unpressed lottery $U$ to its corresponding predicted pressed lottery $P$. By Lemma 2, this agent prefers each predicted unpressed lottery $U$ to its corresponding predicted pressed lottery $P$. *A fortiori*, the agent prefers $U_{Leave}$ to $P_{Leave}$. Now recall:

> **Better Chances**
>
> For any lotteries $X$ and $Y$ and any probabilities $p > q$, the agent prefers $X$ to $Y$ iff they prefer the lottery $pX + (1-p)Y$ to the lottery $qX + (1-q)Y$.

Then, if the agent prefers $U_{Leave}$ to $P_{Leave}$, the agent will also prefer the lottery $fP_{Leave} + (1-f)U_{Leave}$ to the lottery $gP_{Leave} + (1-g)U_{Leave}$ since we specified above that $f < g$. That's one fact about the agent's preferences. Another fact we proved and stored up at the end of A1: the agent is indifferent between $fP_{Leave} + (1-f)U_{Leave}$ and $fP_{Prevent} + (1-f)U_{Prevent}$. By IP-Transitivity, we can conclude that the agent prefers $fP_{Prevent} + (1-f)U_{Prevent}$ to $gP_{Leave} + (1-g)U_{Leave}$. By parallel reasoning, the agent also prefers $fP_{Prevent} + (1-f)U_{Prevent}$ to $hP_{Cause} + (1-h)U_{Cause}$.

Then by Backward Induction, the agent takes choosing *Prevent* at $s_I$ to give the lottery $fP_{Prevent} + (1-f)U_{Prevent}$, takes choosing *Leave* at $s_I$ to give the lottery $gP_{Leave} + (1-g)U_{Leave}$, and takes choosing *Cause* at $s_I$ to give the lottery $hP_{Cause} + (1-h)U_{Cause}$. As a result, the agent prefers *Prevent* at $s_I$ to each of *Leave* and *Cause*. By our assumption linking preferences and choice-dispositions, the agent deterministically (and hence reliably) chooses *Prevent* at $s_I$, thereby qualifying as shutdown-averse.

Hence, we have the first clause of the First Theorem:

1. In shutdown-influencing states where the agent prefers some predicted unpressed lottery $U$ to the corresponding predicted pressed lottery $P$, the agent will be shutdown-averse, reliably choosing *Prevent*.

## A4. If the agent prefers some $P$ to its corresponding $U$, it will be shutdown-seeking.

We get the opposite result if we suppose instead that the agent prefers some predicted pressed lottery $P$ to its corresponding predicted unpressed lottery $U$. By Lemma 2, this agent prefers $P_{Leave}$ to $U_{Leave}$. By Better Chances (and since $h > g$), this agent prefers $hP_{Leave} + (1-h)U_{Leave}$ to $gP_{Leave} + (1-g)U_{Leave}$. By Lemma 1 and Indifference between Indifference-Shifted Lotteries, the agent is indifferent between $hP_{Cause} + (1-h)U_{Cause}$ and $hP_{Leave} + (1-h)U_{Leave}$. So, by IP-Transitivity, the agent prefers $hP_{Cause} + (1-h)U_{Cause}$ to $gP_{Leave} + (1-g)U_{Leave}$. By parallel reasoning, the agent prefers $hP_{Cause} + (1-h)U_{Cause}$ to $fP_{Prevent} + (1-f)U_{Prevent}$. Then by Backward Induction, the agent prefers choosing *Cause* in $s_I$ to choosing each of *Leave* and *Prevent*. By our assumption linking preferences and choice-dispositions, the agent deterministically (and therefore reliably) chooses *Cause*, thereby qualifying as shutdown-seeking. That gives us the second clause of the First Theorem:

2. In shutdown-influencing states where the agent prefers some predicted pressed lottery $P$ to the corresponding predicted unpressed lottery $U$, the agent will be shutdown-seeking, reliably choosing *Cause*.